\documentclass{article} 
\usepackage{iclr2025_conference,times}

\usepackage{graphicx}
\usepackage{xcolor}
\usepackage{algorithm}
\usepackage{algorithmicx}
\usepackage{amsmath}
\usepackage{amssymb}
\usepackage[noend]{algpseudocode}
\usepackage{xspace}
\usepackage{hyperref}
\usepackage{cleveref}
\usepackage{url}
\usepackage{enumitem}

\newif \iffull
\fulltrue

\newcommand{\mathify}[1]{\ensuremath{#1}\xspace}
\newcommand{\privsamp}{\mathify{S_{\mathsf{priv}}}}
\newcommand{\synthdata}{\mathify{S_{\mathsf{synth}}}}
\newcommand{\gemsynth}{\mathify{\mathsf{Gem\_Synth}}}
\newcommand{\numgen}{\mathify{N_{\mathsf{synth}}}}
\newcommand{\eps}{\varepsilon}
\newcommand{\privdist}{\mathify{dist_{\eps}}}
\newcommand{\randapi}{\mathify{\mathsf{Random\_API}}}
\newcommand{\varapi}{\mathify{\mathsf{Variation\_API}}}
\newcommand{\ds}{\mathify{\mathbf{x}}}
\newcommand{\wdist}{\mathify{\mathsf{Wdist}}}
\newcommand{\pmw}{\mathify{\mathsf{PMW^{pub}}}}
\newcommand{\alg}{\mathify{\mathcal{A}}}
\newcommand{\universe}{\ensuremath{\mathcal{U}}\xspace}

\newtheorem{theorem}{Theorem}

\newtheorem{definition}[theorem]{Definition}

\newcommand{\norm}[1]{\left\lVert#1\right\rVert}
\DeclareMathOperator*{\argmin}{arg\,min}

\title{Is API Access to LLMs Useful for Generating Private Synthetic Tabular Data?}

\author{%
  Marika Swanberg\thanks{Work done while an intern at Google Research.}\\
  Boston University\\ 
  \& Google Research\\
  \texttt{marikaswanberg@google.com}\\
  \And
  Ryan McKenna \\
  Google Research\\
  \texttt{mckennar@google.com} \\
  \And
  Edo Roth\\
  Google Research\\
  \texttt{edor@google.com} \\
  \And
  Albert Cheu\\
  Google Research\\
  \texttt{cheu@google.com} \\
  \And
  Peter Kairouz\\
  Google Research\\
  \texttt{kairouz@google.com} \\
}

\iclrfinalcopy 
\begin{document}

\maketitle

\begin{abstract}
    Differentially private (DP) synthetic data is a versatile tool for enabling the analysis of private data. Recent advancements in large language models (LLMs) have inspired a number of algorithm techniques for improving DP synthetic data generation. One family of approaches uses DP finetuning on the foundation model weights; however, the model weights for state-of-the-art models may not be public. In this work we propose two DP synthetic tabular data algorithms that only require API access to the foundation model. We adapt the Private Evolution algorithm \citep{lin2023differentially, xie2024differentially}---which was designed for image and text data---to the tabular data domain. In our extension of Private Evolution, we define a query workload-based distance measure, which may be of independent interest. We propose a family of algorithms that use one-shot API access to LLMs, rather than adaptive queries to the LLM. Our findings reveal that API-access to powerful LLMs does not always improve the quality of DP synthetic data compared to established baselines that operate without such access. We provide insights into the underlying reasons and propose improvements to LLMs that could make them more effective for this application.
\end{abstract}

\section{Introduction}

Synthetic data has long been a ``holy grail'' for performing computations on sensitive data, with the allure of protecting privacy while supporting typical data queries and regular data workflows out-of-the-box. Unfortunately, without a rigorous treatment of privacy, the synthetic dataset may inadvertently reveal information about the sensitive data from which it is derived. 

 Differential privacy (DP) \citep{DworkMNS06, dwork2016calibrating} has emerged as the gold standard for quantifying privacy leakage by algorithms that process sensitive data records from users. At a high level, a (randomized) algorithm satisfies differential privacy if the algorithm's output distribution is not affected very much by a single person's data, regardless of what the other data records are. This ensures that the mechanism's output reveals little about any individual person's data as a result of their participation in the data analysis, even after arbitrary post-processing of the mechanism output. 

Many algorithms have been developed for DP synthetic data, particularly for tabular data \citep{mckenna2022aim,tao2021benchmarking,liu2021iterative,aydore2021differentially,liu2021leveraging,cai2021data,zhang2021privsyn}. With the advancement of large language models (LLMs), a number of recent works propose improved DP synthetic data algorithms that use LLMs trained on public data\footnote{The extent to which data used to train LLMs is considered \emph{public} and compatible with privacy goals is hotly contested~\citep{tramer2022position}. We sidestep this question and assume public models are fair to treat as non-private, but we acknowledge it remains an important question.}. Among these are two broad categories of methods: those which privately finetune a foundation model, and those which only use API access to the foundation model. \citet{sablayrolles2023privately},  \citet{tran2024differentially}, and \citet{afonja2024dp2stageadaptinglanguagemodels} use private finetuning on generative language models to generate private tabular synthetic data, and \citet{kurakin2023harnessing} similarly do private LoRA finetuning on an LLM to generate synthetic text data. Similarly, \citet{ghalebikesabi2023differentially} employ DP finetuning of diffusion models for generating DP synthetic images. 

Despite their power, these DP finetuning methods have significant hurdles. First, finetuning algorithms require white-box access to the model, as the weights need to be directly adjusted. This is a problem because many state-of-the-art models are proprietary, with weights that remain confidential. Only a limited set of researchers are able to even experiment with DP finetuning on such models. Secondly, the resources needed for DP finetuning scales with model dimensionality; time and energy costs quickly become prohibitive. These hurdles motivate alternative ways of using foundation models. In particular, even many proprietary models have a publicly accessible API. 

A series of works in the synthetic image \citep{lin2023differentially} and text \citep{xie2024differentially} domains use only API access to foundation models. The algorithm, Private Evolution, combines adaptive queries to the foundation model with a genetic algorithm to privately generate synthetic image and text data. These methods were further extended to the federated setting by \citet{hou2024pre}. Yet a different approach \citep{amin2024private} uses private prediction combined with other privacy budget saving tricks on the foundation model to generate DP synthetic text.

In light of these recent successes for image and text data, we ask: \textit{Can API access to an LLM improve algorithms for generating DP synthetic \textbf{tabular} data?}

\textit{A priori} it is not obvious why an LLM would be useful at all for generating synthetic tabular data that it was not trained on; however, in initial experiments we found that with descriptive column names, the LLM we used has a reasonable prior over realistic-looking data records. This prior is a powerful source of information we harness in our algorithms.

We design and evaluate two types of DP synthetic tabular generation algorithms that leverage LLM API access. In Section~\ref{sec:private_evolution}, we adapt Private Evolution \citep{lin2023differentially, xie2024differentially} to the tabular domain. A key part of our solution uses a workload-aware family of distance functions, which may be of independent interest, to align the genetic algorithm with the final workload error. In Section~\ref{sec:one_shot} we introduce a new class of private synthetic data algorithms that use one-shot API access to the foundation model. Unlike prior methods, which require adaptive queries to the foundation model or finetuning the model's weights, our method consumes only one (offline) round of queries to the foundation model.
Along the way, we evaluate our two approaches against a number of accuracy baselines to determine whether they advance the state-of-the-art for DP synthetic tabular data.

We evaluated our algorithms with Gemini 1.0 Pro \citep{team2023gemini}, which allowed us to constrain the outputs to structured tabular records. In our evaluations, we find that the proposed methods fail to consistently beat our baselines. Despite this, we think our attempts are instructive to the research community and could inform the development of state-of-the-art methods, especially as foundation models improve. In light of our findings, we share our key take-aways:

 \paragraph{The role of data domain.} The state-of-the-art for DP synthetic data generation is highly domain specific. In particular, DP tabular synthetic data has been very well-studied compared to image and text, so the state of the art for tabular data is much harder to improve on. 
 Additionally, prior work on Private Evolution relies on public image and text embeddings to measure the fidelity of the synthetic data, but similar embeddings do not exist for tabular data. Our workload-aware distance function in Section~\ref{sec:private_evolution} is one substitute, but surely other solutions exist as well.

\paragraph{The importance of appropriate baselines.} In the tabular data domain, there is no single algorithm that dominates on all datasets, query workloads, and privacy budgets. Any new algorithm in this area requires extensive comparison to the handful of algorithms that dominate the state-of-the-art, as well as naive baselines. In Section~\ref{sec:one_shot}, we show that combining Gemini-generated data with JAM \citep{fuentes2024joint} outperforms all other methods; however, in testing other baselines we find that this holds regardless of the public data we give JAM. Without this naive baseline, we would have reached a false conclusion that the Gemini-generated data was improving the state-of-the-art.

\iffull
\section{Preliminaries}
We begin by presenting the definition of differential privacy, which is a constraint on an algorithm \alg that processes a dataset $\ds = (\ds_1, \ldots, \ds_n)$ of user records, one per user. Two datasets are called \emph{neighbors} if they differ on one person's record. At a high level, differential privacy requires that for any pair of neighboring datasets, the algorithm's output distributions are similar when run on each dataset.

\begin{definition}[Differential Privacy~\citep{DworkMNS06, dwork2016calibrating}]\label{def:DP} A randomized algorithm $\alg: \universe^{n} \rightarrow \mathcal{Y}$ is {\em $\eps$-differentially private} if for every pair of neighboring datasets $\ds, \ds'\in \universe^n$, and for all outputs $y \in \mathcal{Y}$,
 \begin{equation*}
    \Pr[\alg(\ds) =y] \leq e^\eps \cdot \Pr[\alg(\ds') = y] + \delta,
 \end{equation*}
 where the probability is taken over the internal coins of \alg.
 \end{definition}
 
 The differential privacy guarantee is parameterized by $\eps>0$, where algorithms with lower values have less privacy leakage and higher values of epsilon denote more privacy leakage from the algorithm's output. DP gives a worst-case guarantee (over the algorithm's inputs and outputs) on how much information an algorithm leaks about its input.

\subsection{Prior Work}\label{sec:prior_work}

\paragraph{GAN-based methods for DP synthetic data}

Many prior works have proposed synthetic data mechanisms based on generative adversarial networks.  See \cite{yang2024tabular} for a nice survey of these and other approaches.  These mechanisms generally work by fitting the parameters of the model via DP-SGD, and then using the model to generate synthetic data after training.  These techniques are typically best suited for unstructured data like images or text.

\paragraph{Marginal-based methods for DP synthetic data}

Many mechanisms for DP synthetic data generation work by adding noise to low-dimensional marginals of the data distribution \cite{mckenna2021winning,mckenna2022aim,cai2021data,aydore2021differentially,fuentes2024joint,vietri2022private,liu2021iterative,liu2021leveraging,zhang2021privsyn}.  Some mechanisms in this space are also designed to leverage public data when it's available \cite{fuentes2024joint,liu2021iterative,liu2021leveraging}.  Benchmarks have confirmed these approaches work very well in tabular data settings \cite{tao2021benchmarking}.

\fi
\section{Adapting Private Evolution to Tabular Data}\label{sec:private_evolution}

We adapt Private Evolution (PE) \citep{lin2023differentially, xie2024differentially} to the tabular data domain. Private Evolution works in rounds, by maintaining a set of \emph{candidates} $S_t$ generated by the foundation model and using a distance function together with a differentially private histogram to have each private record individually vote for candidates. The best performing synthetic candidates become part of an \emph{elite set} for that round $S'_t$; at the end of each round, the foundation model is prompted to generate more examples similar to the elite set, which then become the new candidates $S_{t+1}$.

The first set of candidates are populated by a \randapi, which prompts the model to generate some prespecified number of initial candidates adhering to the column names and datatypes of the private dataset. Each subsequent set of candidates are generated via the \varapi which takes the current elite set of candidates and prompts the model to generate some number of additional candidates that are similar.

\begin{algorithm}
    \caption{Private Evolution \citep{lin2023differentially, xie2024differentially}}
    \label{alg:private_evolution}
    \hspace*{\algorithmicindent} \textbf{Input:} Private samples \privsamp, Number of iterations $T$, Number of generated samples \numgen, Distance function $\privdist(\cdot, \cdot)$, Noise multiplier $\sigma$\\
    \hspace*{\algorithmicindent} \textbf{Output:} Synthetic data \synthdata
    \begin{algorithmic}[1] 
        \State $S_1 \gets \randapi(2\cdot\numgen)$
        \For{$t=1$ to $T$}
            \State $H = []$ \Comment{Initialize histogram over $S_t$}
            \For{$x_{priv}\in \privsamp$}
            \State $i = \arg\min_{j\in[n]} \privdist(x_{priv}, S_t)$ \Comment{Compute closest synthetic candidate}
            \State $H[i] = H[i] + 1$
            \EndFor
            \State $H \gets H + \mathcal{N}(0, \sigma I_{2 \cdot \numgen})$ \Comment{Add noise to ensure DP}
            \State $H \gets \max(0, H)$ \Comment{Post-process element-wise}
            \State $\mathcal{P}_t \gets H/sum(H)$ \Comment{Compute empirical distribution on $S_t$}
            \State $S_t' \gets $ draw \numgen samples with replacement from $\mathcal{P}_t$
            \State $S_{t+1} \gets \varapi(S'_t)$
        \EndFor
        \State \Return $S_T$
    \end{algorithmic}
\end{algorithm}

\subsection{Workload-Aware Distance Function}
Prior methods that applied Private Evolution to image \citep{lin2023differentially} and text \citep{xie2024differentially} data used public text and image embeddings, respectively, to measure the distance between candidate synthetic examples and the private examples. Choosing a sensible distance function for tabular records is less straightforward: public tabular embeddings (if they exist) likely wouldn't capture the features of unseen data, simple $\ell_p$ distance fails to account for differences in scale among columns. 

Instead, we derive a workload-aware distance function. A private synthetic dataset is typically optimized for and evaluated on a particular \emph{workload} of (linear) queries $W = \{q_1, \ldots, q_k\}$. The workload error is typically some $\ell_p$ variation on:

$$\mathsf{WError}(\privsamp, \synthdata) = \sum_{i\in [k]} |q_i(\privsamp) - q_i(\synthdata)|.$$ 

Note that workload error is a function of pairs of datasests; however, the distance function we require is a function of pairs of individual records. We unpack the workload error further: assuming the queries are \emph{linear}, then they correspond to a sum over a predicate on data records $q_i(\ds) = \sum_{j\in[n]} \psi_i(x_j)$. Thus, for the given predicates $\psi = (\psi_1, \ldots, \psi_k)$ corresponding to the queries in $W$, we will define the workload-aware distance function between a private record and synthetic candidate: 
$$
\wdist_\psi(x, c) = \sum_{i\in k} |\psi_i(x) - \psi_i(c)|.
$$
A dataset of synthetic candidates with low workload-aware distance will have low workload error compared to the private data.

\subsection{Experimental results}

We evaluate our adapted private evolution algorithm on a modified version of NYC Taxi and Limousine Commission data using Gemini 1.0 Pro. We use data from January 2024, to to avoid data contamination between the public and private evaluation data \citep{geminiModelCard}.

\begin{figure}
    \centering
    \includegraphics[width=\textwidth]{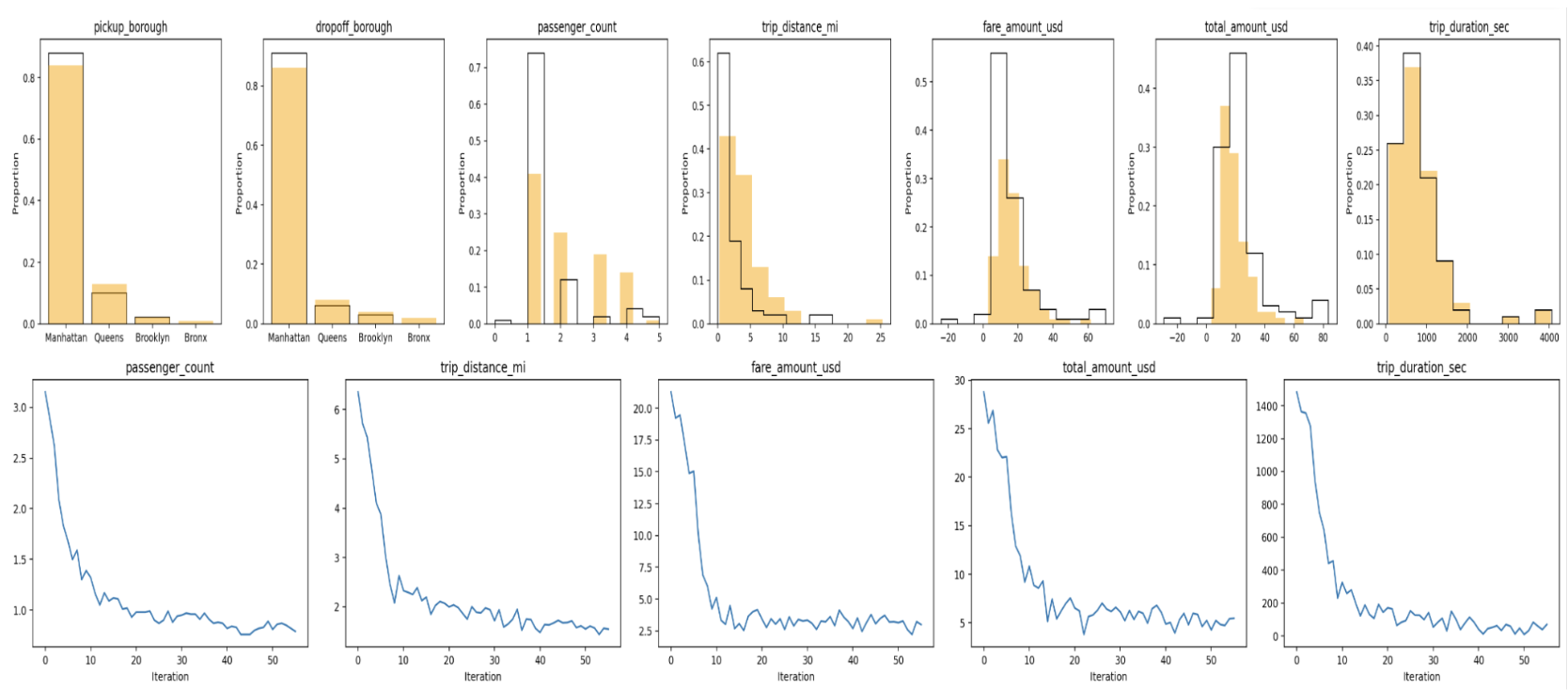}
    \caption{\textbf{Top} 1-way marginals on private (outlined) and synthetic (yellow) data. \textbf{Bottom} workload error of synthetic data over time for Private Evolution with $\eps = \infty$.}
    \label{fig:pe_converge}
\end{figure}

For our workload, we use a scaled $\ell_1$-distance on numerical variables for each combination of categorical variables. We rescale the numerical variables to account for different value ranges---for example, trip distance (in miles) versus trip duration (in seconds). As an initial experiment, we ran the algorithm without any privacy constraints, and we found that the workload error converges and the 1-way marginals of the synthetic data converges to the 1-way marginals on the private data as well (see Figure~\ref{fig:pe_converge}). It's worth noting that, depending on the expressiveness of the foundation model, it's not a given that the PE algorithm would converge even without privacy.  

We then ran the same experiments but with a DP histogram instead of a nonprivate one. We experimented with various hyperparameters: how many initial random examples to use, how many iterations to use, how to split the budget across iterations, etc. We found that using increasing budget across runs worked better than an even or decreasing budget; additionally, having more candidates relative to \numgen worked best, and finally, using fewer iterations worked best---in fact, using \emph{only a single iteration} was the optimal setting we could find. 

With differential privacy, the private evolution algorithm failed to beat two simple baselines: \emph{independent} which privately computes all 1-way marginals and samples data from the product over the private marginals, and \emph{DP workload} which directly computes the workload queries with DP, without generating any synthetic data. These two baselines are not the only two which we'd hope to beat, rather, they're the bare minimum. Moreover, querying any large foundation model hundreds of times is relatively slow. 

\paragraph{What we learned} The observation that the workload-aware private evolution algorithm performs best with \emph{one shot} data generation implies that: \emph{whatever marginal gains we get from iterating multiple times, they are outweighed by the privacy cost of composing over iterations.} Additionally, while PE was developed for image and text domains where finetuning a foundation model is the main alternative for DP synthetic data, there is a vast literature on algorithms for private synthetic \emph{tabular} data that do not require access to generative models. These lessons paved the way for our second attempt, which proved to be more successful, though still did not beat the current state-of-the-art.

\section{Using Gemini-Generated Records as Public Data}\label{sec:one_shot}

\iffull
As overviewed in \Cref{sec:prior_work}, there 
\else
There
\fi
is a substantial body of work on DP synthetic tabular data. Some state-of-the-art algorithms within this space make use of \emph{public data} to improve the accuracy or efficiency of the algorithm on private data; however, for many applications such public data may not be available in the format required\footnote{This is especially true for algorithms that assume the public dataset has the same (or substantially overlapping) columns, or even is distributed similarly to the private data.}, as is discussed in-depth in \cite{liu2021iterative}[Section 6.1]. Our second approach uses Gemini generated data in lieu of this public data. 

\subsection{Approach overview}
 Using Gemini's structured output functionality, we prompt Gemini to generate data records with a response schema matching the column names and datatypes of our private dataset. Importantly, none of the private records influence the prompts to Gemini---only the column names and datatypes do. This data generation occurs ``offline'' and in one shot with no loss of privacy budget. We call this dataset \gemsynth. Later, we plug this synthetic public dataset into various DP synthetic tabular algorithms that use public data as well as the private data. 

One major benefit of the one-shot nature of this method (rather than querying Gemini interactively in a loop) is we can generate \emph{many} synthetic public data records and reuse the same generated records when trying different approaches. This is not possible when the records are generated adaptively as in Private Evolution. Thus, this method takes advantage of our observations about PE. 
We begin with a high-level overview of how public data is incorporated into two DP synthetic data algorithms. For both, we consider what happens when we use Gemini-generated tabular data as the public data source for these algorithms.

\paragraph{\pmw \cite{liu2021leveraging}} The \pmw algorithm is an improvement of MWEM~\citep{hardt2012simple}, which we will not discuss in detail. The basic idea is to use public data to initialize the generating distribution over synthetic records and iteratively refine this distribution to reduce the workload error. The public records reduce the number of iterations required, by providing a ``warm start'' for the synthetic data distribution, along with reducing the data domain over which the distribution is estimated.

A key sub-routine of \pmw is to estimate a distribution that approximately matches some noisy statistics.  Specifically, let $Q$ denote a collection of linear queries and let $\tilde{y} = Q(D) + \xi = \sum_{x \in D} Q(x) + \xi $ be the noisy answers to those queries on the sensitive data.  \pmw finds a distribution supported on the ``public'' data \gemsynth, and finds the weights to assign to each public record to minimize the $\ell_2$ squared error to the noisy observations.

$$ w^* = \argmin_{w \in \mathbb{R}_+} \norm{ \sum_{x \in \gemsynth} w_x Q(x) - \tilde{y} }_2^2 $$ 

When the public records are sufficiently representative, this method can work quite well.  However, with small or unrepresentative public datasets, this method may not find a good distribution even in the absence of noise.

\paragraph{Gemini inference} We use the Gemini-generated records as the public records for the subroutine of \pmw, calling this ``Gemini inference'', setting $Q$ to be the query workload.

\paragraph{MST modified to take public data} The standard MST algorithm \citep{mckenna2021winning} has three phases: selecting marginal queries, measuring the marginals with DP, and lastly using Private-PGM to post-process the noisy marginals and generate a synthetic dataset. We modify the final step (generation), replacing Private-PGM with the subroutine from \pmw that utilizes \gemsynth.   This method differs from the Gemini inference approach primarily in how the queries $Q$ are selected. 

\paragraph{JAM} The JAM-PGM mechanism \citep{fuentes2024joint} was developed for marginal queries, and utilizes public data in a different manner.  It privately decides whether to measure each marginal query on the public data or the private data in order to minimize the overall workload error. This mechanism has the benefit that it can utilize public data that is accurate on some, but not necessarily all, marginals. We run this mechanism as-is, using \gemsynth as the public data.

\subsection{Baselines for Comparison}
Because there are a wealth of methods for generating private synthetic data with and without public data, we have a fair number of baselines that we need to compare any new methods to. A number of works that privately finetune foundation models for tabular data omit comparisons to state-of-the-art methods for generating DP tabular data, so it is unclear if they outperform existing approaches. 

We study two baselines that require no privacy. First, we consider an \textbf{in-distribution public dataset:} publically data drawn from the same distribution as the private data. This is essentially the lowest error we could hope for, up to sampling error; however, in-distribution public data is usually not available. Second, we consider using the \textbf{Gemini data with no DP} to answer workload queries. 

Next, we consider a number of baselines that do not require public data. 
\begin{itemize}
    \item \textbf{DP workload}: we compute the queries directly using DP.
    \item \textbf{Independent baseline}: privately compute the 1-way marginals and sample records from the corresponding product distribution over marginals.
    \item \textbf{MST algorithm}: a tabular synthetic data algorithm that does not use public data.
    \item \textbf{\pmw with uniform data} and \textbf{JAM with uniform data}: using data that is drawn uniformly from the domain as public data for these algorithms.
\end{itemize}

\subsection{Experimental setup}

Our private dataset is UCI Adult \citep{adult_2}. Using this structured output constraint, we sample Gemini with top-$k$=1 and temperature=1 to generate 131,000 records in \gemsynth. We use 2-way marginals as our query workload to evaluate the fidelity of the DP synthetic data.

\subsection{Results}
\begin{figure}
    \centering
    \includegraphics[width=0.4\textwidth]{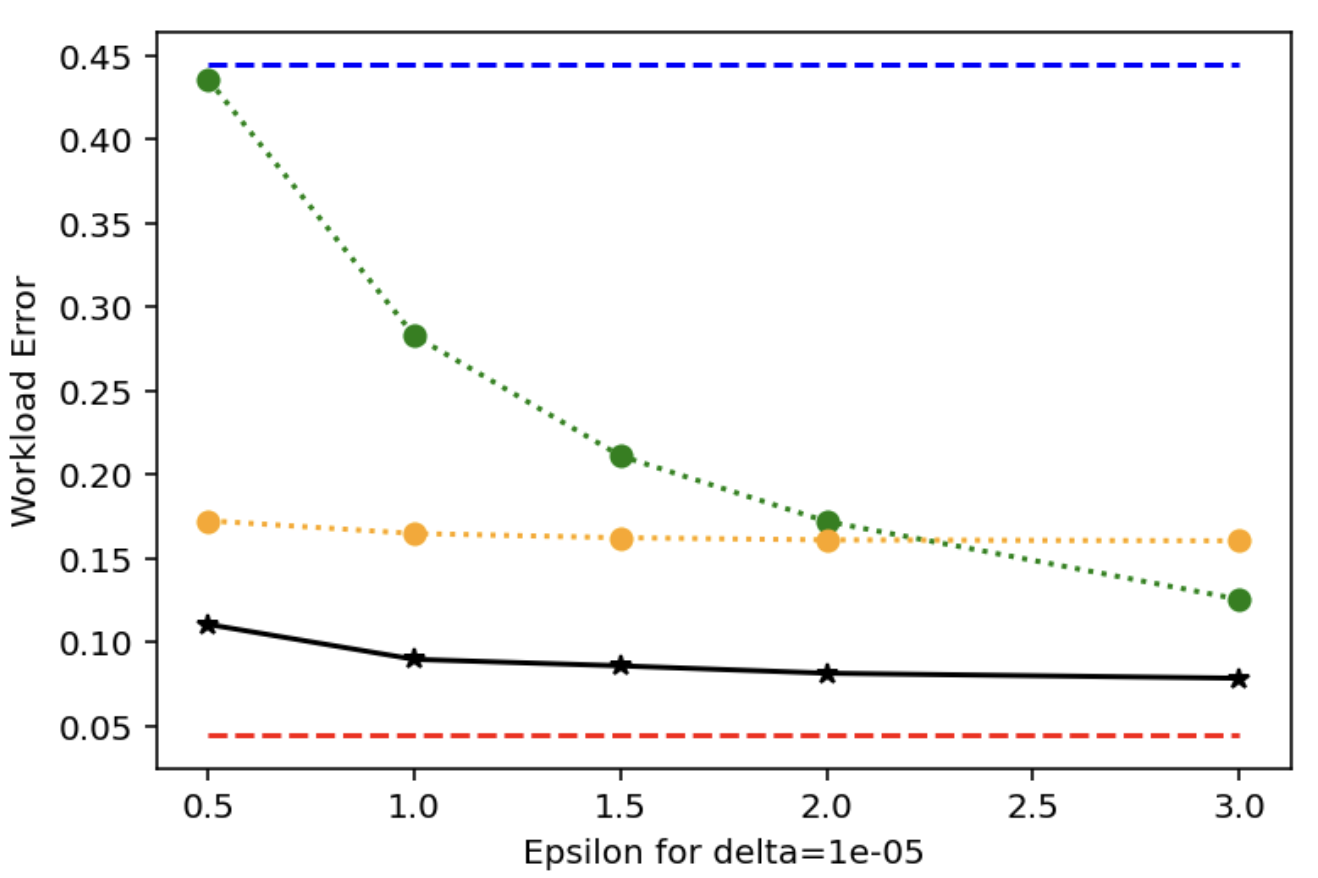}
    \includegraphics[width=0.59\textwidth]{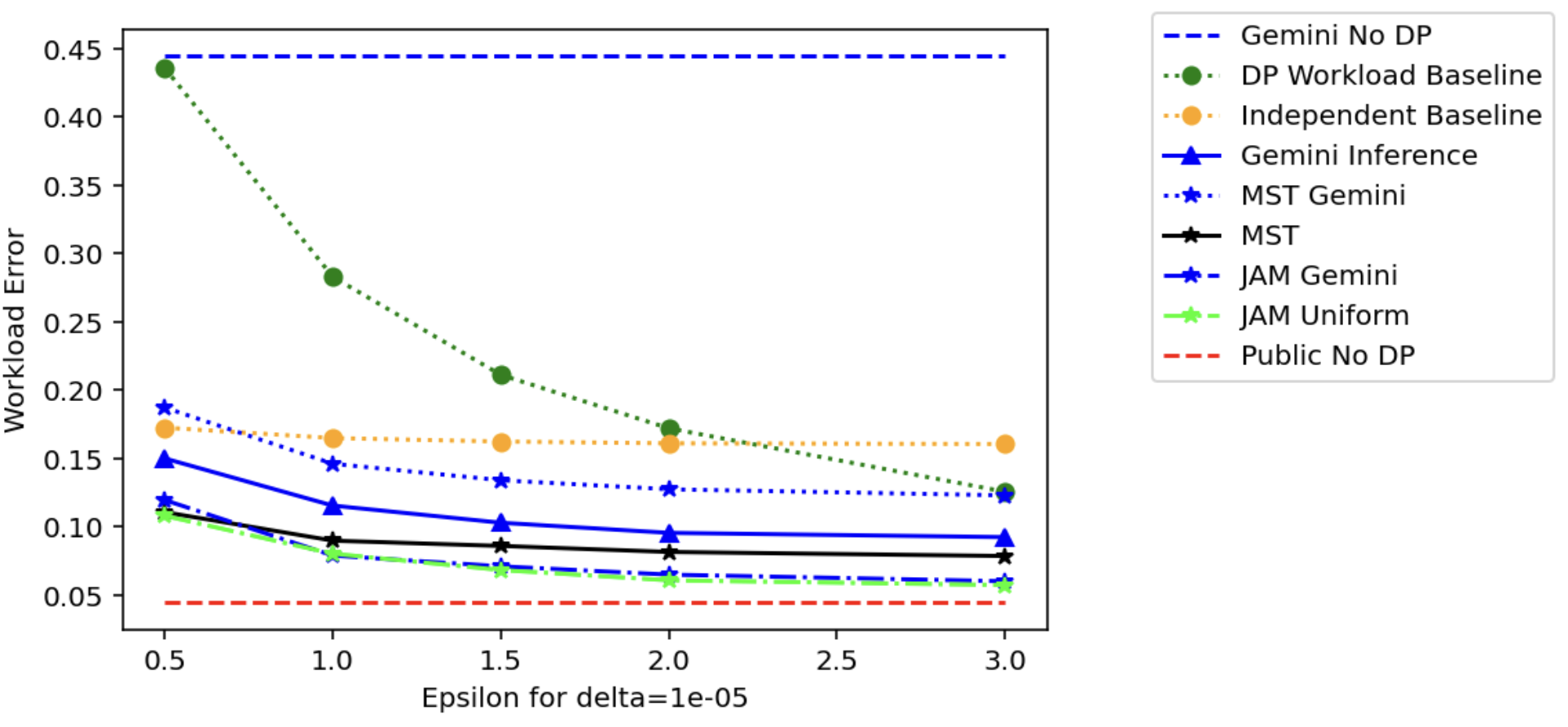}
    \caption{\textbf{(Left)} Workload error for baseline methods for generating tabular synthetic data without use of Gemini. \textbf{(Right)} Workload error for baseline methods and our one-shot methods that use API access to Gemini.}
    \label{fig:results}
\end{figure}

Figure~\ref{fig:results} (Left) shows the workload error versus epsilon for the baseline methods discussed. Among these methods, MST achieves the lowest workload error (except the in-distribution public dataset which is our unachievable ``best-case baseline''). Figure~\ref{fig:results} (Right) shows all of the results for the baseline methods in addition to the methods that use \gemsynth. Notice that JAM with \gemsynth performs best overall; however, JAM performs equally well with uniform data. This is because JAM is simply using the private data to compute answers to the queries rather than utilizing the public data. Thus, JAM with \gemsynth is not better than the state-of-the-art methods on this dataset and query workload. In general, \gemsynth may capture 1-way marginals on the data better than uniform, however it is generally inaccurate on $k$-way marginals.

\section{Conclusion and Future Work}

We evaluated two methods for incorporating API access to Gemini for generating DP synthetic tabular data. While our methods did not beat state-of-the-art methods, this work motivates a number of future directions. First, as foundation models continue to improve, combining our methods with better models (e.g. models trained on more tabular data) could potentially improve the final accuracy, especially if the models are trained specifically for the tabular setting. Additionally, because Gemini uses word embeddings, perhaps doing some finetuning on publicly available tabular data could improve the quality of the Gemini-generated tabular records fed into our one-shot method. Lastly, perhaps there are ways to achieve better accuracy by combining Private Evolution and our one-shot approach. Using foundation models for DP synthetic data generation is still a very new area of research, with many avenues for improvements and breakthroughs.

\bibliography{bib.bib}
\bibliographystyle{iclr2025_conference}

\iffull
\else

\newpage
\appendix

\fi

\end{document}